\def\year{2022}\relax
\documentclass[letterpaper]{article} 
\usepackage{aaai22}  
\usepackage{times}  
\usepackage{helvet}  
\usepackage{courier}  
\usepackage[hyphens]{url}  
\usepackage{graphicx} 
\usepackage{natbib}  
\usepackage{caption} 
\DeclareCaptionStyle{ruled}{labelfont=normalfont,labelsep=colon,strut=off} 
\usepackage{algorithm}
\usepackage{algorithmic}

\usepackage{amssymb}
\usepackage{amsmath}
\usepackage{subfigure}

\usepackage{multirow}
\usepackage{graphicx}
\usepackage{epstopdf}

\usepackage{color}

\usepackage{newfloat}
\usepackage{listings}
\floatstyle{ruled}
\newfloat{listing}{tb}{lst}{}
\floatname{listing}{Listing}
\title{DetarNet: Decoupling Translation and Rotation by Siamese Network for Point Cloud Registration}
\author{
	Zhi Chen, Fan Yang, Wenbing Tao\textsuperscript{}\thanks{Corresponding author.}\\
}
\affiliations{
	National Key Laboratory of Science and Technology on Multispectral Information Processing \\
	School of Artifical Intelligence and Automation, Huazhong University of Science and Technology, China\\
	\{z\_chen,  fanyang, wenbingtao\}@hust.edu.cn 
}

\usepackage{bibentry}
\definecolor{Blue}{RGB}{0,0,255}

\begin{document}

\maketitle

\begin{abstract}

Point cloud registration is a fundamental step for many tasks. In this paper, we propose a neural network named DetarNet to decouple the translation $t$ and rotation $R$, so as to overcome the performance degradation due to their mutual interference in point cloud registration.
First, a Siamese Network based Progressive and Coherent Feature Drift (PCFD) module is proposed to align the source and target points in high-dimensional feature space, and accurately recover translation from the alignment process. Then we propose a Consensus Encoding Unit (CEU) to construct more distinguishable features for a set of putative correspondences. After that, a Spatial and Channel Attention (SCA) block is adopted to build a classification network for finding good correspondences. Finally, the rotation is obtained by Singular Value Decomposition (SVD). In this way, the proposed network decouples the estimation of translation and rotation, resulting in better performance for both of them. 
Experimental results demonstrate that the proposed DetarNet improves registration performance on both indoor and outdoor scenes. Our code will be
available in \url{https://github.com/ZhiChen902/DetarNet}.

\end{abstract}

\section{Introduction}

\noindent Point cloud registration is one of the fundamental problems in computer vision, which is widely applied to 3D reconstruction, robotics, autonomous driving and medical tasks. It aims to establish correspondences between two point clouds, and estimate the rigid transformation (translation $t$ and rotation $R$). The most commonly used way is first establishing coarse correspondences, and then recovering rigid transformation. The main challenge is that there always exist wrong correspondences (outliers). Although some methods attempt to generate more accurate correspondences through hand-crafted \cite{rusu2009fast,rusu2008persistent} or deep-learning technique based descriptors \cite{zhou2018learning,yew20183dfeat,choy2019fully}, it is hard to be totally outlier-free when dealing with complicated scenarios. Thus, it is worth studying how to better perform point cloud registration in the scenarios when the initial correspondences contain outliers.

\begin{figure}[t]
	\centering
	\includegraphics[width=1\columnwidth]{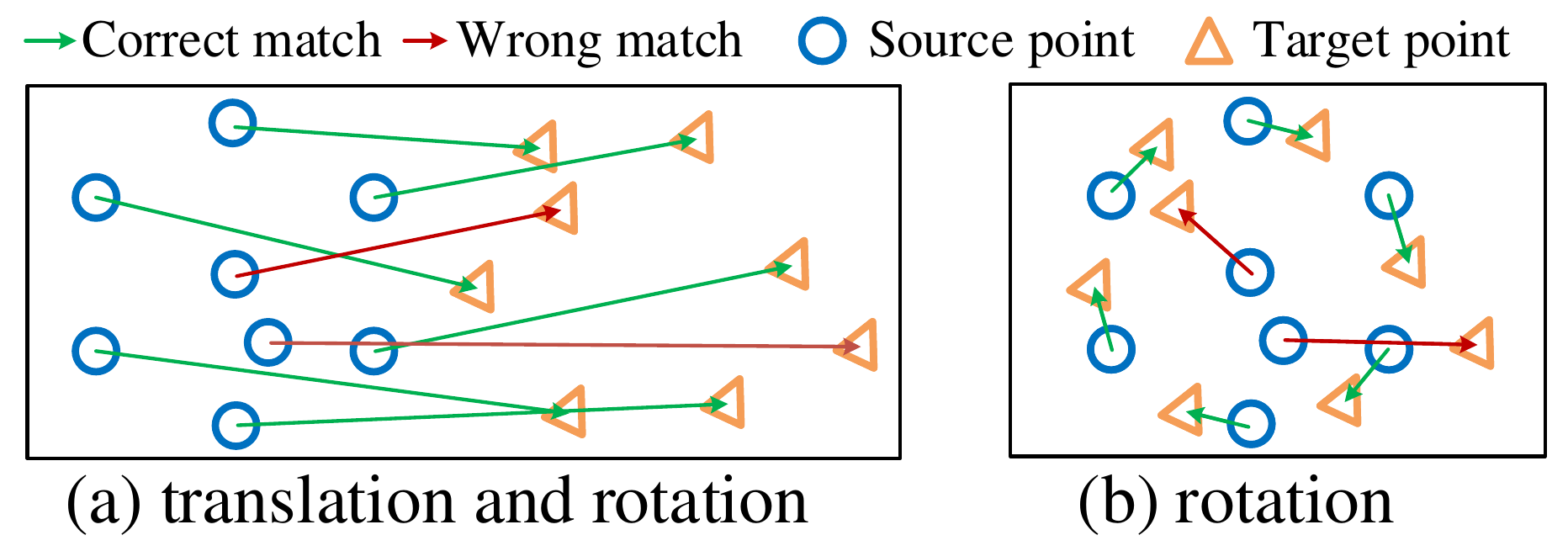}
	\caption{(a): A toy example shows the challenge when translation and rotation are coupled. When accurately estimating translation in advance and there only exists rotation as shown in (b), the consensus between inliers is easier to be mined.}\label{CEU}
\end{figure}


Recently, some methods have studied how to use a classification neural networks \cite{pais20203dregnet,choy2020deep,bai2021pointdsc,lee2021deep} to find correct 3D correspondences between two point clouds, and then estimate the translation $t$ and rotation $R$ by weighted Singular Value Decomposition (SVD) \cite{choy2020deep}.
The core of these methods is to learn the consensus \cite{bai2021pointdsc} of correct correspondences (inliers). Each correspondence can be abstracted as an arrow between a pair of points, as shown in Fig. \ref{CEU}. The consensus of inliers is that the length and direction of inliers are satisfied with some consistency. As shown in Fig. \ref{CEU} (a), due to the coupling of translation and rotation, the consistency between inliers is difficult to be mined. However, as illustrated in Fig. 1 (b), if we can first eliminate the translation $t$ and only the rotation is left, it is much easier for us to find out the correct correspondences. Inspired by this observation, we expect to decouple the whole rigid transformation into separate translation and rotation estimation.


Considering the non-linearity of the rotation space \cite{peng2019pvnet,li2018deepim}, it is more feasible to first recover the translation $t$ because it is linear and easy to handle.
However, decoupling $t$ and solving it accurately in advance is still challenging in two aspects: 
1) It is hard for traditional geometric optimization methods to only recover $t$ without considering $R$. These methods usually need to jointly optimize $t$ and $R$. 
Although centroid alignment \cite{arun1987least} can yield a rough $t$, it can only be used to assist the optimization of the whole rigid transformation due to the existence of outliers. 
2) Although deep learning networks have made remarkable progress in point clouds registration, translation transformation is still hard to be separately modeled in the neural networks while excluding the influence of $R$. Most of the methods try to regress both $t$ and $R$ \cite{pais20203dregnet,aoki2019pointnetlk} together.

Based on the above analysis, we propose a Siamese Network based Progressive and Coherent Feature Drift (PCFD) module to decouple $t$ from the whole transformation and solve it accurately. PCFD module converts the registration into an alignment process of two point clouds in high-dimensional feature space. First, the features of the two point clouds are respectively extracted by a Siamese Network with shared parameters. Then a global feature offset is learned by establishing global interaction between the two point clouds. The global feature offset forces the source points to move towards the target points coherently as a group to preserve the topological structure of point sets. Thus, the transformation is explicitly encoded by the alignment process, which is named as Coherent Feature Drift (CFD) operation. The whole PCFD module is composed of multiple CFD operations, which progressively align the source and target points to obtain the optimal estimation of $t$.

The formulation of PCFD module has three advantages:
1) CFD operations explicitly encode the transformation by the global features in the network. Thus, the coupling between $R$ and $t$ can be disentangled by introducing the supervision on the middle layers. We supervise the alignment process by using only the ground truth translation $t_{gt}$, so that the global features tend to encode the translation transformation. 
2) When putative correspondences are given, previous methods usually \cite{choy2020deep,pais20203dregnet,bai2021pointdsc} concatenate the two points of a correspondence and form a virtual point to process together. Different from them, our PCFD module adopts a Siamese Network to retain the features of two point clouds. 
Since the two point clouds are handled respectively, it is easier to establish interaction between them, which benefits the regression for transformation.
3) The network adopts a progressive alignment approach to regress $t$ and gradually eliminates $t$ by using a multi-layer CFD operation. The multiple layers of CFD constitute an iterative optimization structure, so $t$ can be more accurately estimated.

Since we obtain the accurate estimation of $t$, the correspondences between two point clouds are more obvious and easier to be decided, as shown in Fig. \ref{CEU} (b). Then we follow the previous works \cite{moo2018learning,pais20203dregnet} and build a correspondence classification network to prune outliers. Specifically, a Consensus Encoding Unit (CEU) is proposed to remove $t$ when encoding the consensus to make the feature more distinguishable. It combines the spatial and feature consistency items as the feature for each correspondence. Furthermore, we design a Spatial and Channel Attention (SCA) block for the construction of classification network. It simplifies the current spatial attention module \cite{sun2020acne,chen2021cascade} and combines it with an instance-unique channel attention. Thus, the network can capture more complex context to better find the consensus of inliers. 
Finally, according to the established correspondences, $R$ is obtained by Singular Value Decomposition (SVD) \cite{arun1987least}.

The above modules are integrated into an end-to-end registration network named DetarNet. In a nutshell, our main contributions are threefold: \textbf{1.} We propose a Progressive and Coherent Feature Drift (PCFD) module to gradually align the source and target points in feature space. With this process, the $t$ vector can be accurately recovered. \textbf{2.} We propose a Consensus Encoding Unit (CEU) to construct a feature for each correspondence and a Spatial and Channel Attention (SCA) block to find correct correspondences. They can establish accurate matches for $R$ estimation. \textbf{3.} The above modules are integrated to build decoupling solutions for $R$ and $t$. Experiments show that the proposed network achieves state-of-the-art performance on both indoor and outdoor datasets.
	




\section{Related Works}
\textbf{Feature-Based 3D Matching.} A common way to establish correspondences between 3D point clouds is by extracting local descriptors. Traditional hand-crafted descriptors are usually generated by extracting the local information, such as histograms of spatial coordinates \cite{frome2004recognizing,johnson1999using,tombari2010unique} and geometric attributes \cite{chen20073d,salti2014shot}. 
Some other methods \cite{rusu2008persistent,rusu2009fast} aim to design rotation invariant descriptors.
Recently, deep learning techniques are explored to learn 3D descriptors. Many of these methods \cite{su2015multi,zhou2018learning,zeng20173dmatch,deng2018ppfnet} take the point cloud patches as input to learn local features. Some other methods \cite{choy20194d,choy2019fully,yew20183dfeat,bai2020d3feat,huang2021predator} use point clouds as input to generate dense feature descriptors on point clouds. Although the methods above can usually establish good initial correspondences, it is hard to be totally outlier-free in the application. Our method is to address the challenge of registration when there are outliers in the correspondences.

\begin{figure*}[t]
	\centering
	\includegraphics[width=2\columnwidth]{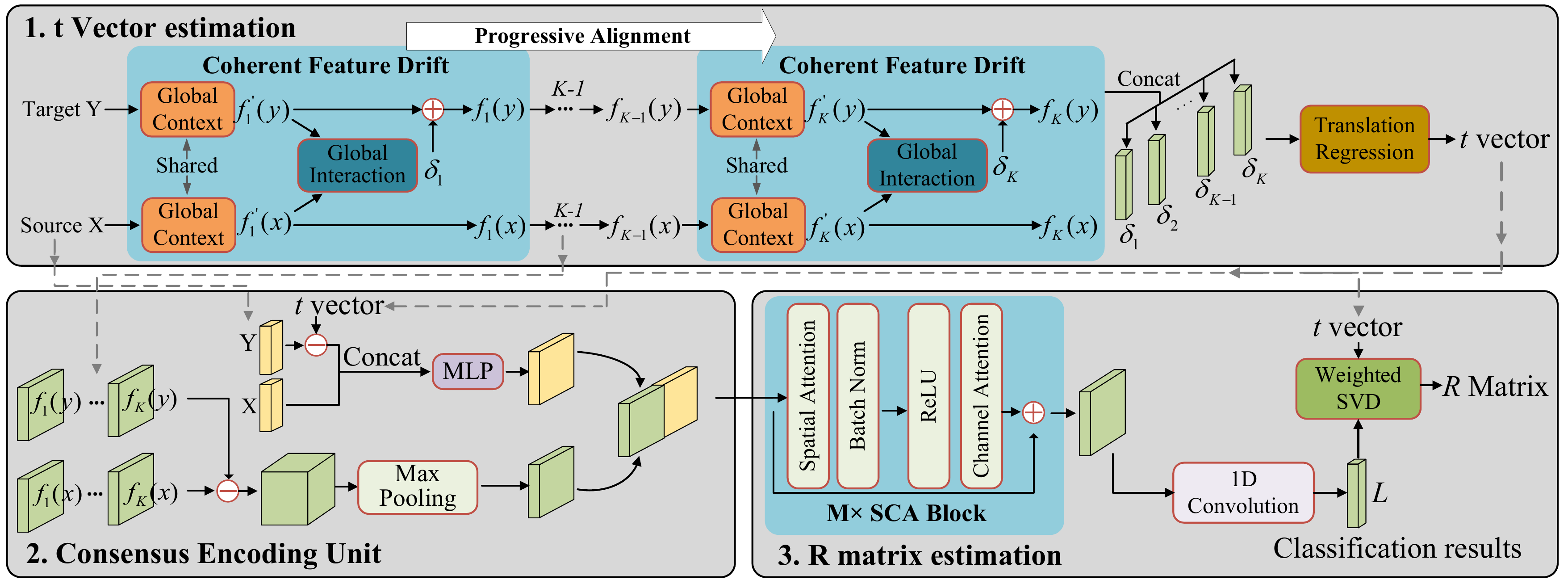}
	\caption{
		Overview of our network. 1. A Progressive and Coherent Feature Drift (PCFD) module progressively aligns the source and target points in feature space, and recovers $t$ vector from the alignment process. 
		2. An Consensus Encoding Unit (CEU) constructs feature for each correspondence by combining the spatial and feature consistency. 3. Several Spatial and Channel Attention (SCA) blocks are adopted to find correct correspondences, and followed with weighted SVD for estimating $R$ matrix.}\label{pipeline}
\end{figure*}

\noindent\textbf{Outlier Removal.} Given a putative correspondence set that contains outliers, one can use outlier removal methods to remove outliers. The most widely used method is the RANdom SAmple Consensus (RANSAC) \cite{fischler1981random}, and its variants \cite{chum2005matching,fragoso2013evsac,brahmachari2009blogs,goshen2008balanced}.Recently, some methods start adopting deep learning techniques to find good 2D-2D correspondences. The CN-Net \cite{moo2018learning} proposes a Context Normalization (CN) operation for finding correct correspondences. 
Later works \cite{plotz2018neural,zhang2019learning,zhao2019nm,sun2020acne,brachmann2019neural,liu2021learnable} aim to capture more context to enhance the performance.
Besides, recent attempts try to use deep learning networks for finding 3D correspondences, such as 3DReg-Net \cite{pais20203dregnet}, DGR-Net \cite{choy2020deep} and PointDSC \cite{bai2021pointdsc}. Our work aims to better find correct 3D correspondences and align point clouds through decoupling translation and rotation transformations.

\noindent\textbf{Pose Estimation.} 
Pose estimation is the final goal of rigid 3D point registration, i.e., estimating a rigid transformation to align point clouds. Besl and McKay \cite{besl1992method} propose the iterative closest point (ICP) algorithm to align point cloud through iteratively establishing point correspondence and performing least squares optimization. The variants of ICP \cite{rusinkiewicz2001efficient,segal2009generalized,bouaziz2013sparse} are proposed to address the challenges existing in ICP, such as efficiency, partiality and sparsity. Recently, some methods adopt end-to-end frameworks for directly estimating the rigid transformation between point clouds. Deep Closest Point (DCP-Net) \cite{wang2019deep} uses deep global features to form correspondences and estimate relative pose. Later works \cite{yew2020rpm,wang2019prnet} aim to address the problem of partial visibility to further improve the performance of registration.

\section{Methods}
Given two point clouds to be registered: $X=\{x_{i} \in \mathbb{R}^{3} \; | \; i = 1, ..., N_{x}\}$ and $Y = \{x_{j} \in \mathbb{R}^{3} \; | \; j = 1, ..., N_{y}\}$, we first form $N$ pairs of correspondences as follows:
\begin{equation}\label{input}
	C=\begin{bmatrix}
		x_{1} &  x_{2} & ... & x_{n} \\ 
		y_{1} &  y_{2} & ... & y_{n}
	\end{bmatrix} \in \mathbb{R}^{2 \times N \times 3},
\end{equation}
where  $x_i$ and $y_i$ ($1 < i < N$) are a pair of matched points.
These putative correspondences are established by extracting local descriptors and matching. Limited by the distinctiveness
of descriptors, many of these correspondences are wrong (outliers) .
The goal of the network is to recover the rigid transformation from these noisy correspondences.
It takes the coordinates of these correspondences as input,
and outputs the probability of being correct (inliers) for each correspondence and rigid transformation as follows:
\begin{equation}\label{input}
	t, \; R, \;  L  = \Phi(C); t \in {\mathbb{R}}^{3}, R \in {\mathbb{R}}^{3 \times 3}, L \in {\mathbb{R}}^{N \times 1}, 
\end{equation}
where $\Phi(\cdot)$ is the network with trained parameters. $t$ and $R$ are the estimated translation and rotation respectively. $L$ is the logit value of each correspondence being inlier. 
In this paper, we propose a decoupling solution for the $t$ and $R$. 
The pipeline of our method is shown in Fig. \ref{pipeline}. We will explain the details of each module in the following sections.


\subsection{Translation Estimation}
\label{Siamese}
\textbf{Progressive and Coherent Feature Drift.}
The PCFD module transforms point cloud registration into a process of coherently moving source points to target points. Since deep neural network can extract more informative feature for each point, we convert the coherent drift operation to high-dimensional feature space. 
As shown in Fig. \ref{pipeline}, the PCFD module is composed of $K$ Coherent Feature Drift (CFD) operations. Each CFD tries to align the features of source and target points generated by the previous CFD layer, so it is a progressive process.

Specifically, the CFD first encodes feature for each point in a Siamese architecture.  We use the CN Block \cite{moo2018learning}, which is a variant of PointNet \cite{qi2017pointnet}, for encoding global context. Formally, let $f_{l-1}(x)$ and $f_{l-1}(y)$ be the output of $l-1$ layer, then $l$-th CFD encodes the features as follows:
\begin{equation}
	\begin{aligned}
		f_{l}^{'}(x)={\rm CN}(f_{l-1}(x)), f_{l}^{'}(y) = {\rm CN}(f_{l-1}(y)),
	\end{aligned}
\end{equation}
where $f_{l}^{'}(x)$ and $f_{l}^{'}(y)$ are the extracted features for source and target points. 
Note that the CN operations for source and target points are  parameter-shared. In this way, the feature difference between a pair of correspondences is completely caused by the rigid transformation between them. 
Then we perform coherent drift by moving the features of source points to target points. A core of coherent drift is to force the source points to move coherently as a group to preserve the topological structure of the point
sets \cite{myronenko2010point}. To ensure coherent constraints, a global feature offset ($\delta_{l}, 1 \leq l \leq K$) shared by all the source points is learned the $l$-th CFD. After that, we hold the target points and move the source points are moved by the $\delta_{l}$ to the target points.
\begin{equation}
	\begin{aligned}
		f_{l}^{}(x_{i})=f_{l}^{'}(x) + \delta_{l}, f_{l}(y_{i})=f_{l}^{'}(y_{i}); 1 \leq i \leq N,
	\end{aligned}
\end{equation}
where $x_{i}$ and $y_{i}$ are the $i$-th point in the source and target points. $f_{l}(x_{i})$ and $f_{l}^{}(y_{i})$ are the output feature.

\begin{figure}[t]
	\centering
	\includegraphics[width=1\columnwidth]{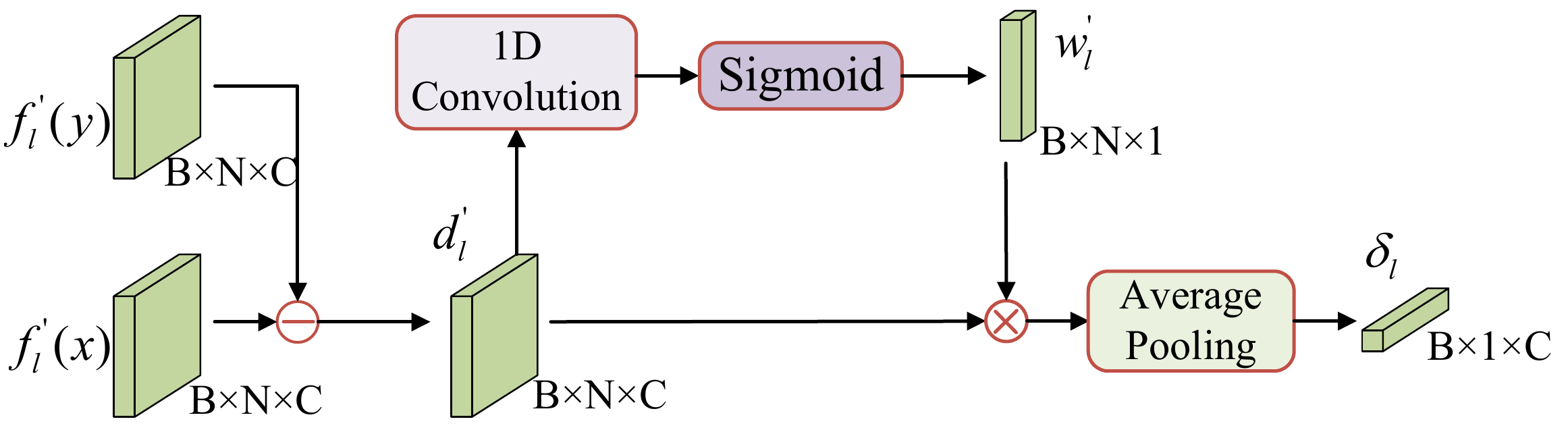}
	\caption{The global interaction operation.}\label{GlobalInteraction}
\end{figure}

An important issue of CFD is how to learn the global feature offset $\delta_{l}$. Each $\delta_{l}$ needs to make the source points gradually approach the target points in the feature space.
In the CFD operation, a global interaction is adopted to learn it as in Fig. \ref{GlobalInteraction}. 
It first computes the feature difference between the feature of source and target points, as follows:
\begin{equation}
	\begin{aligned}
		d_{l}^{'} = f_{l}^{'}(y) - f_{l}^{'}(x).
	\end{aligned}
\end{equation}
As mentioned before, $x_{i}$ and $y_{i}$ are a pair of putative correspondences. So $d_{l}^{'}$ is the feature offset between putative correspondences. We then learn a weight ($w_{l}^{'}$) for each correspondence by a convolution and sigmoid function:
\begin{equation}
	\begin{aligned}
		w_{l}^{'} = sigmoid(Conv(d_{l}^{'})) 
	\end{aligned}
\end{equation}
Finally, we use an average pooling to integrate the feature difference of all correspondences to produce the global feature offset $\delta_{l}$. The 1D convolution plays two roles in the learning of $\delta_{l}$: 1) Since there are many outliers in the putative correspondences, the weights produced by the convolution and sigmoid function is expected to suppress the outliers. 2) 
There are learnable parameters in the convolution operation to increase the flexibility of global interaction.

\noindent\textbf{Supervising Drift Process.} 
In order to reduce the interference of $R$ for better encoding the translation transformation, we introduce supervision in the middle of the network. Supervision is applied to the global feature offset $\delta_{l}$ ($1 < l < N$).
Specifically, we first add up all the previous offsets, then use a 1D convolution to regress a temporary $t_{l} \in \mathbb{R}^{3}$ vector as follows:
\begin{equation}
	\begin{aligned}
		t_{l} = Conv(Sum(\delta_{1}, ..., \delta_{l-1}, \delta_{l})).
	\end{aligned}
\end{equation}
Then a drift loss is used to supervise all layers of $t_{l}$:
\begin{equation}
	\begin{aligned}\label{drift_loss}
		\mathcal{L}_{align} = \frac{1}{K}\sum_{l=1}^{K} \{\frac{1}{N}\sum_{i=1}^{N}m_{i} \cdot \rho(y_{i}, R_{gt}x_{i} + t_{l}) \},
	\end{aligned}
\end{equation}
where $\rho(., .)$ is the distance metric function. $R_{gt}$ is the ground truth rotation. $m_{i}$ is the mask for correspondence $i$. $m_{i}$ is set to 1 if the ground-truth of correspondence $i$ is inliers. Otherwise, it will be set to be 0. 
It is a semi-alignment loss that uses the estimated translation $t_{l}$ in $l$-th layer and ground-truth rotation $R_{gt}$ to align the two point cloud and penalizes the alignment error.
Thus, it expects all alignment to approximate the accurate $t$ vector.

\noindent\textbf{Translation Regression.}
As mentioned before, every time a CFD is performed, the source point cloud is globally aligned to the target point cloud by the global feature offset, and $\delta_{l}$ encodes the alignment process. We can naturally solve the $t$ matrix by integrating the offsets of all layers. We concatenate all of the $\delta_{l}$ ($1 \leq l \leq K)$ and then adopt a 1D convolution to regress $t$ vector.

\subsection{Consensus Encoding Unit}
An important issue for finding correct correspondences from putative correspondences is to mine consensus of inliers \cite{pais20203dregnet,choy2020deep,bai2021pointdsc}, so that outliers can be distinguished from inliers.
As introduced in Introduction Section and Fig. \ref{CEU}, it becomes easier to mine the consensus when removing translation $t$ and remaining only rotation $R$ between two point clouds. Since our PCFD module can regress $t$ vector in advance, the Consensus Encoding Unit (CEU) tries to remove translation for better encoding consensus. The architecture of CEU is shown in Fig. \ref{pipeline}, it combines the consensuses in coordinate and feature space.

For the consensus in coordinate space, it is intuitive to remove the $t$ vector. 
We subtract the estimated $t$ vector from the target point cloud, so that the translation $t$ between the source point cloud and the target point cloud is removed. 
Then the coordinate offset between the source point cloud and the target point cloud is followed with a 1D convolution to be as a feature. 
Meanwhile, CEU also tries to mine feature consistency between the correct correspondences. It utilizes the feature produced by the previous PCFD module to construct feature for correspondence to integrate more information. As introduced before, in PCFD module, source points are aligned to target points in feature space. By introducing the supervision of the intermediate layer, the $t$ transformation between the source points and the target points is removed in feature space. Thus, we use the feature difference of these layers to construct the feature for the correspondence, which can encode consensus without translation. In order to make full use of the context of shallow and deep networks, all the layers of PCFD module are used to form a multi-layer correlation feature. Then a max-pooling, which performs the best with other choices based on our experiments, is adopted to integrate multi-layer context. The consensus feature in coordinate and feature space is combined by a concatenation operation.


\begin{figure}[t]
	\centering
	\includegraphics[width=1\columnwidth]{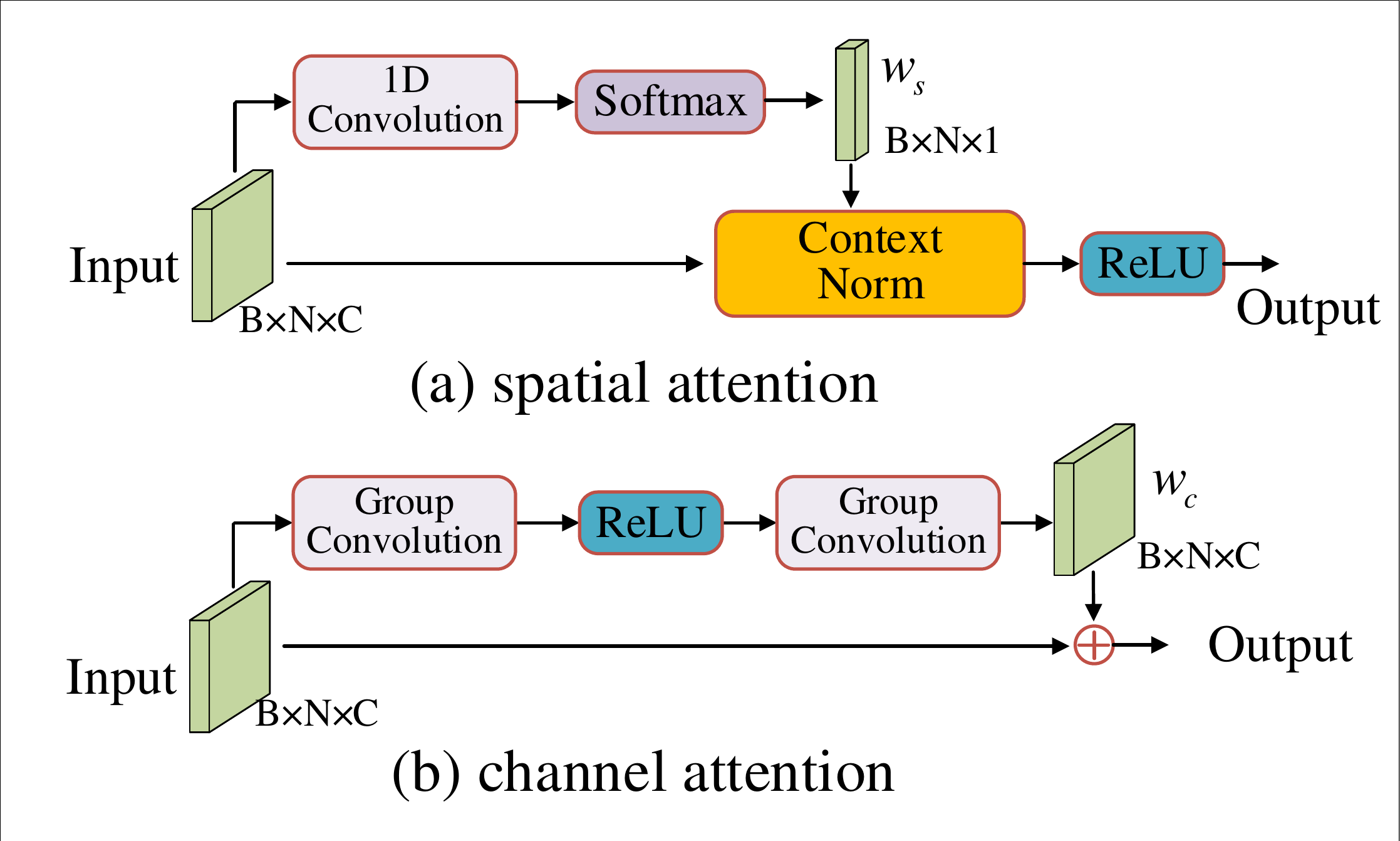}
	\caption{
		The Spatial Attention and Channel Attention in SCA block.}\label{SCA}
\end{figure}


\subsection{Rotation Estimation}
After Consensus Encoding Unit constructs a feature for each correspondence, a classification network is adopted to finding correct correspondences, and followed with weighted SVD to recover $R$ matrix.

\noindent\textbf{Classification Network.}
As shown in Fig. \ref{pipeline}, the classification network is composed of $M$ Spatial and Channel Attention (SCA) Blocks. Each SCA block integrates a spatial attention, batch normalization \cite{ioffe2015batch}, ReLU and a channel attention in a ResNet architecture. The spatial attention is already proposed for finding 2D-2D correspondences \cite{sun2020acne,chen2021cascade} operation. They integrate local, global and prior information to learn weight for learning global context to ignore outliers. In this paper, since the CEU can construct more distinguishable features, we simplified the spatial attention to reduce the parameters of the network as Fig. \ref{SCA} (a). For the input feature, it first learns a weight vector ($w_{s} \in \mathbb{R}^{B \times N \times 1}$) by means of a 1D convolution and Softmax function. Then the weight vector is utilized as guidance to perform a weighted context normalization \cite{moo2018learning} for encoding global context. The weight vector is to allow outliers to be ignored when performing context normalization.
Meanwhile, interdependencies between feature channels are proved to be helpful for feature learning \cite{hu2018squeeze}. So we also introduce a channel attention operation as Fig. \ref{SCA} (b). Different from the commonly used SE-Net, we use instance-unique channel attention. It learns an independent weight vector for each instance instead of a shared one. Thus, it can capture more complex channel information for each correspondence. In order to reduce network computation for learning the instance-unique weight map, the group convolution \cite{cohen2016group} is used instead of the regular one.



\noindent\textbf{Weighted SVD.}
We use weighted SVD \cite{choy2020deep}, which reformulates the traditional SVD \cite{arun1987least} into a weighted version, to recover $R$ matrix.
Specifically, $x_{i}$ and $y_{i}$ ($1 \leq i \leq N $) are the points in source and target point clouds respectively. 
We first use the estimated $t$ vector to process the target points:
\begin{equation}
	\begin{aligned}
		y_{i}^{'} = y_{i} - t, 1 \leq i \leq N.
	\end{aligned}
\end{equation}
Then a weighted matrix $H$ for SVD is computed as follows:
\begin{equation}
	\begin{aligned}
		H = \sum_{i \in \mathcal{I}}{w_{i} x_{i} {y_{i}^{'}}^{T}}, H \in \mathbb{R}^{3 \times 3},
	\end{aligned}
\end{equation}
where the weight $w_{i}$ is computed by the logit value of classification as follows:
\begin{equation}
	\begin{aligned}
		w_{i} = {\rm tanh}({\rm ReLU}(L_{i})), 1 \leq i \leq N,
	\end{aligned}
\end{equation}
Finally, $R$ can be obtained by performing SVD on $H$ matrix as follows:
\begin{equation}
	\begin{aligned}
		R = U diag(1,1,det(UV^{T}))V^{T}, H = U \sum V^{T}.
	\end{aligned}
\end{equation}

%
\subsection{Loss Function}
We formulate our training objective as a combination of four types of loss functions, including translation loss ($l_{trans}$), classification loss ($l_{cls}$), alignment loss ($l_{align}$) and drift loss ($l_{drift}$) as follows:
\begin{equation}
	\begin{aligned}\label{loss_function}
		loss = \lambda_{1}l_{trans} + \lambda_{2}l_{cls} + \lambda_{3}l_{align} + \lambda_{4}l_{drift}
	\end{aligned}
\end{equation}
$l_{trans}$ is the L2 loss between the ground-truth and estimated $t$ vector. $l_{cls}$ is the cross entropy loss. $l_{align}$ penalizes the wrong alignment between correct correspondences as follows:
\begin{equation}\label{alignment_loss}
	\begin{aligned}
		l_{align}=\frac{1}{N}\sum_{i=1}^{N}m_{i} \cdot \rho(y_{i}, Rx_{i} + t),
	\end{aligned}
\end{equation} 
where $\rho(., .)$ is Euclidean distance. $t$ and $R$ are the estimated translation and rotation transformation. $N$ is the number of correspondences. $m_{i}$ is also the mask for correspondence $i$ to label inliers, as introduced in Eq. \ref{drift_loss}.  $l_{drift}$ is to supervise the middle layer as Eq. \ref{drift_loss}.  

\begin{figure}[t]
	\centering
	\includegraphics[width=1\columnwidth]{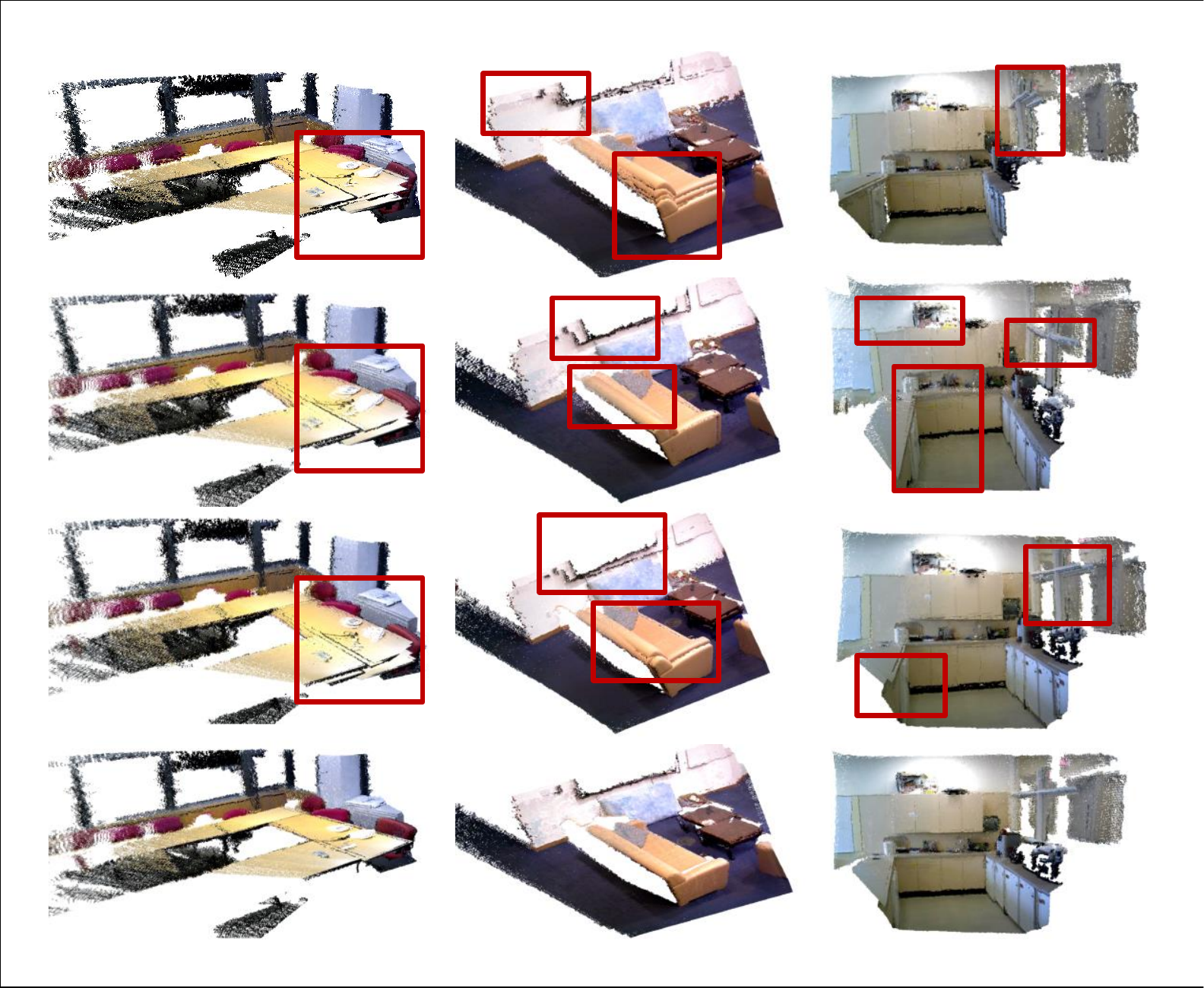}
	\caption{The visualized alignment results of four different methods. From top row to bottom: 3DRegNet \cite{pais20203dregnet}, RANSAC \cite{fischler1981random}, FGR \cite{zhou2016fast} and ours. The alignment areas with large errors are marked with red boxes. Our method achieves the best alignment result among these methods.} \label{visualization}
\end{figure}

\section{Experiments}

\subsection{Experimental Setup}

\textbf{Outdoor Dataset.} We use the KITTI \cite{geiger2012we} odometry dataset, which contains 11 outdoor driving scenarios of points clouds. We follow the splitting way of previous works \cite{bai2020d3feat,choy2019fully} and use scenario 0 to 5 for training, 6 to 7 for validation and 8 to 10 for testing. Then for each point cloud, we construct 30cm voxel grid to downsample the point cloud \cite{choy2019fully}. 
\begin{table*}[t]
	\centering%
	\caption{Quantitative results on the KITTI, SUN3D and 7Scenes Datasets. The mean rotation error (MRE), mean translation error (MTE), mAP and recall under the threshold of ($5^{\circ}$, 15 cm) are reported.} \label{tab:overallPreformance}%
	\small
	\begin{tabular}{c|cccc||cccc||cccc|| c}
		
		\hline
		%
		
		& \multicolumn{4}{c||}{KITTI} & \multicolumn{4}{c||}{SUN3D} & \multicolumn{4}{c||}{7Scenes (Generalization )} & \\ \hline 
		&  MRE  & MTE   & $\rm mAP$ & $\rm recall$  &  MRE  & MTE   & $\rm mAP$ & $\rm recall$ &  MRE  & MTE   & $\rm mAP$ & $\rm recall$ & Time  \\ \hline %
		
		ICP & 1.208 & 90.21 & 0.54 & 1.41 & 6.178 & 15.40 & 19.6 & 49.7  & 5.504 & 12.44 & 27.8 & 50.3 & 0.28\\
		RANSAC & 0.759 & 29.65 & 43.1 & 80.9 & 3.580 & 15.16 & 43.9 & 80.5  & 2.107 & 12.41 & 32.2 & 68.5 & 2.79\\
		
		GCRANSAC & 0.152 & 70.41 & 1.89 & 3.12 & 1.920 & 9.672 & 37.6 & 75.6 & 1.946 & 10.43 & 30.7 & 78.1 & 0.82\\
		
		FGR  & 0.298 & 12.13 & 31.4 & 72.0  & 2.895 & 10.85 & 39.2 & 73.6  & 2.913 & 13.52 & 31.5 & 66.3 & 0.32\\
		
		DGR & 0.157 & 9.773 & 41.6 & 82.0 & 2.239 & 9.663 & 41.3 & 82.8 & 2.166 & 13.54 & 30.0 & 63.4 & 0.76\\ 
		PointLK & 5.352 & 43.84 & 4.01 &  7.25 & 7.732 & 27.64 & 17.2 & 32.0  & 26.49 & 32.37 & 5.12 & 9.61 & 0.13\\ 
		
		PointDSC & 0.152 & 8.966 & 46.9 & \textbf{91.5} & 1.913 & 7.283 & 50.1 & 89.7 & \textbf{1.902} & 11.31 & 38.6 & 78.4 & 0.09 \\
		3DRegNet  & 0.752 & 31.62 & 12.3 & 28.7 & 2.889 & 13.13 & 31.2 & 68.6 & 6.424 & 15.21 & 26.7 & 58.2 & 0.03 \\ 
		
		Ours & \textbf{0.148} & \textbf{8.126} & \textbf{48.1} & 88.1 & \textbf{1.840} & \textbf{5.317} & \textbf{56.3} & \textbf{93.1} & 2.011 & \textbf{7.739} & \textbf{42.7} & \textbf{84.9} & 0.04\\ \hline
	\end{tabular}
	
\end{table*}

\noindent\textbf{Indoor Datasets.} We use the SUN3D dataset \cite{xiao2013sun3d} to generate the dataset for training and testing. Sun3D is composed of 268 sequences of RGBD videos. We randomly select 115 sequences for training and validation, and 20 sequences for testing. 
For each video sequence, we first subsample the videos by a factor of 10. Then for each frame, we recover the point cloud by depth map, and construct 5cm voxel grid to downsample the point cloud \cite{zhou2018open3d}.
The 7scenes \cite{shotton2013scene} dataset contains 46 RGBD sequences under various camera motion statuses, we follow the official split to use the 18 sequences of them as test dataset. It is adopted for generalization experiments.

\noindent\textbf{Data Processing.}
Following 3DReg-Net \cite{pais20203dregnet}, we use FPFH descriptors \cite{rusu2009fast} to generate 2560 pairs of correspondences between adjacent frames as input. Then we generate the ground-truth rotation and translation according to the offered camera pose of each frame and label the correspondences as inlier/outlier (1 refers inliers and 0 refers outliers) by a predefined distance threshold.

\noindent\textbf{Evaluation Metrics.}
For a pair of point clouds, we evaluate the results by computing the errors between the estimated and ground-truth rigid transformation. The errors of rotation (RE) are evaluated by the isotropic error \cite{ma2012invitation}.
The errors of translation (TE) are evaluated by the $L2$ error \cite{choy2020deep}. 
For the whole test dataset, we first report the mean of rotation (MRE) and translation (MTE) errors.
Then, given an error threshold of $R$ and $t$, we can determine whether each estimated pose is accurate or not. We build a normalized cumulative precision curve of pose estimation in the whole test set. After that, we use ($5^{\circ}$, 15 cm) as threshold to figure the recall \cite{choy2020deep} and the area under the curve as mean average precision (mAP) \cite{moo2018learning}.


\noindent\textbf{Implementation Details.}
In the PCFD module, we use 10 layers of CFD to progressively align the two point clouds ($K$ = 10 in Fig. \ref{pipeline}). In the classification module, 4 SCA blocks are utilized to build classification network (M = 4 in Fig. \ref{pipeline}). The number of channels in all layers of the network is set to 128. During training, $\lambda_{1}$, $\lambda_{2}$, $\lambda_{3}$ and $\lambda_{4}$ in loss function (Eq. \ref{loss_function}) are set to 2, 1, 1 and 0.05 respectively. The network is trained by Adam optimizer \cite{kingma2014adam} with a learning rate being $\rm 10^{-3}$ and batch size being 16. All the experiments are conducted on a machine with an INTEL Xeon E5-2620 CPU and a single NVIDIA GTX1080Ti. 
For time-consuming, to do a fair comparison for all the methods, all computation timings are obtained using CPU.

\subsection{Comparison to Other Baselines} 
We compare our method with other baselines, including ICP \cite{besl1992method}, FGR \cite{zhou2016fast}, RANSAC \cite{fischler1981random}, GCRANSAC \cite{barath2018graph}, DGR \cite{choy2020deep}, PointLK \cite{aoki2019pointnetlk}, PointDSC \cite{bai2021pointdsc} and 3DRegNet \cite{pais20203dregnet}. ICP, FGR, RANSAC and GCRANSAC are classical methods while DGR, PointLK, PointDSC and 3DRegNet are learning based methods. All the learning based networks are retrained with the same dataset. For ICP, RANSAC and FGR, we use the version Open3D implemented, while the released codes are adopted for other methods.
We present the quantitative results on the KITTI, SUN3D and 7Scenes Datasets.
The results on 7Scenes are obtained by the model trained on SUN3D dataset as generalization experiments.
As shown in Tab. \ref{tab:overallPreformance}, the recall and mAP of our method are higher than other methods. It shows the overall performance of our methods. More specifically, the $t$ error of our method is much smaller than other methods, especially on indoor scenes. The $t$ error of our method is 7.81cm and 7.47cm smaller than that of our baseline network (3DRegNet) on SUN3D and 7Scenes datasets. It implies that the proposed Progressive and Coherent Feature Drift (PCFD) module can boost the performance of $t$ estimation. For time-consuming, since our network can output the results  without repeated sampling as RANSAC \cite{fischler1981random} and post-processing, it is faster than other methods except for 3DRegNet.
Finally, in order to visually demonstrate the registration performance, we present the visualized alignment results in Fig. \ref{visualization}. We select multiple point clouds and calculate the relative pose between each point cloud and its neighbor. Then we transform these point clouds into the same coordinate frame. The results of 3DRegNet \cite{pais20203dregnet}, RANSAC \cite{fischler1981random}, FGR \cite{zhou2016fast} are presented as comparison. Our method achieves the best alignment results with fewer errors.

\subsection{Registration Robustness} 
\begin{figure}[t]
	\centering
	\includegraphics[width=1\columnwidth]{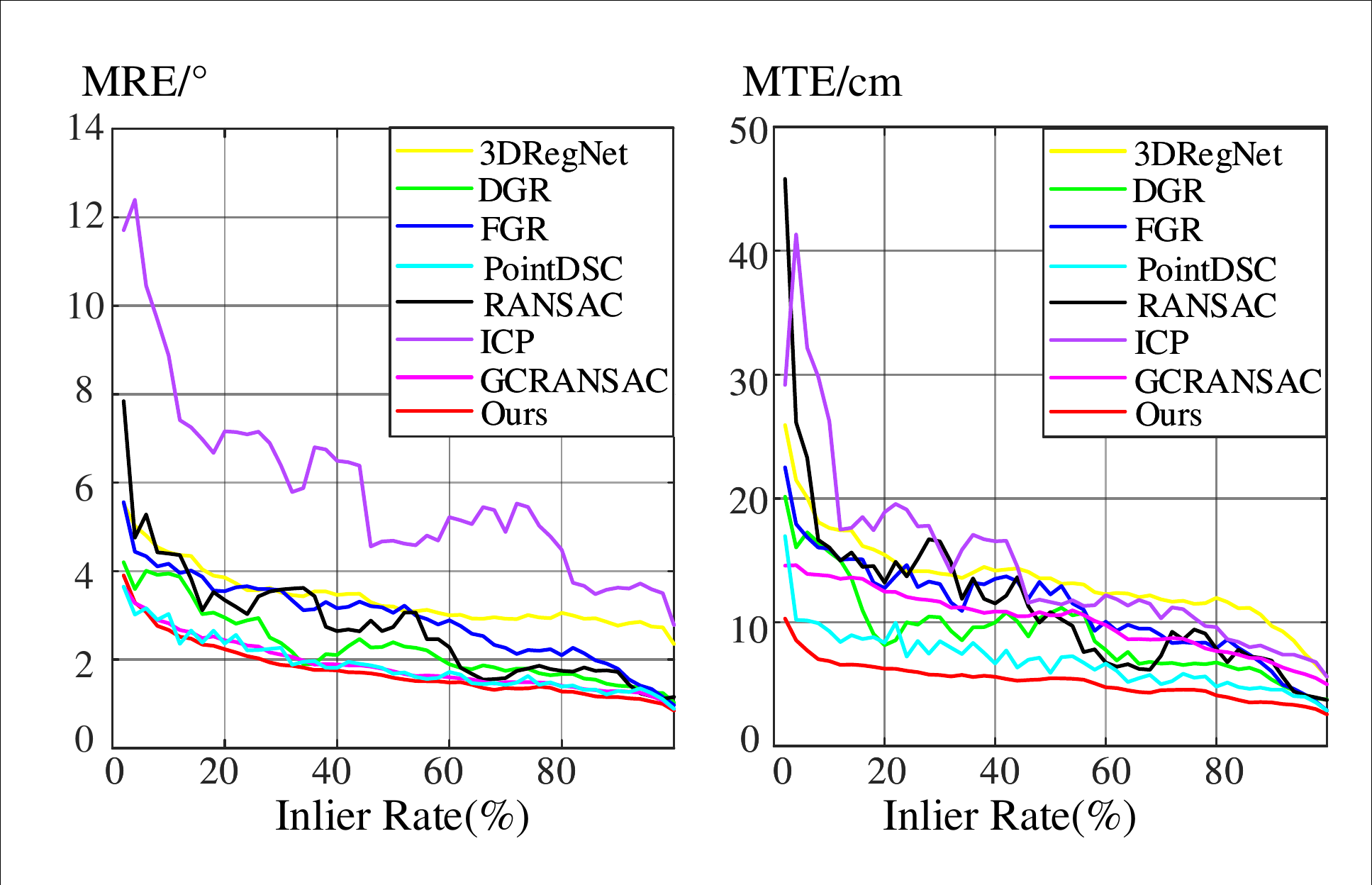}
	\caption{The error curves of $R$ (MRE) and $t$ (MTE) under different inlier ratios of initial correspondences on SUN3D dataset.} \label{error_outliers}
\end{figure}

So far, we have demonstrated the overall performance of the proposed network. In order to further analyze the registration robustness anti-noise, we test the performance under the scenarios with the different inlier ratios of initial correspondence set. Specifically, we divide the test set of SUN3D dataset into several subsets according to the inlier ratio, and respectively compute the mean errors of $R$ (MRE) and $t$ (MTE) estimation at each inlier ratio, as shown in Fig. \ref{error_outliers}. As we can see, our method has obtained results with smaller errors under the scenarios of different inlier ratios for both $R$ and $t$. It demonstrates that our method is robust to outliers. 
Besides, when the inlier ratio changes, the error range of our method is also smaller, which shows that the performance of our method is relatively stable.

\subsection{Method Analysis} 
In this section, we will analyze our method in detail. Expect the previously introduced four evaluation metrics, we also report the classification accuracy (Acc in Tab. \ref{SVDorReg} and \ref{Ablation}) for better understanding the effect of each module.
 
\noindent\textbf{Regression or SVD - Tab. \ref{SVDorReg}.}
In our network, $t$ is estimated by regression while $R$ is solved by SVD.  
We discuss these two estimation heads for $t$ and $R$. The regression and weighted SVD are adopted as the estimation heads for $t$ and $R$, respectively. Through permutation and combination, we can generate four alternatives. For each alternative, we use the proposed network as the backbone for feature extraction.
By analyzing these four groups of control experiments, we can get the following observations: 1) When the estimation heads of $R$ are consistent, direct regression will get better results than the SVD method for $t$ estimation. 
When the estimated head of $t$ is the same, the $R$ result obtained by SVD is better. This proves that regressing $t$ in advance, which adopted in our method is a good choice.
2) We further compare the results of the 2-th group and the 4-th group (ours). We can find that our method obtains more accurate $t$. Meanwhile, although the 2-th and the 4-th group use the same estimation head for $R$ estimation, the 4-th group still achieves better results for $R$ estimation. This can be explained as our method can estimate and remove $t$ in advance before finding the correspondence. Thus, there exists only rotation between correspondence, which helps the classification of inliers and outliers. In fact, the 4-th group dose achieve a better classification result than that of the 2-th group. The above results prove the effectiveness of the idea of decoupling $t$ and $R$.

\begin{table}[t]
	\renewcommand\tabcolsep{2.6pt}%
	\centering%
	\caption{The registration result of using different estimation heads.} \label{SVDorReg}%
	\small
	
	\begin{tabular}{c|cc|ccccc}
		\hline 
		
		Tag & $t$  & $R$ & Acc &  MRE & MTE  & $\rm mAP$ & $\rm recall$ \\ \hline
		
		1 & Reg & Reg & 65.7 & 2.45 & 6.22 & 49.7 & 86.5 \\
		2 & SVD & SVD & 69.1 & 2.29 & 9.17 & 43.2 & 80.0 \\
		3 & SVD & Reg & 60.2 & 2.69 & 9.01 & 35.7 & 62.3 \\
		4 & Reg & SVD & \textbf{72.9} & \textbf{1.84} & \textbf{5.32} & \textbf{56.3} & \textbf{93.1} \\

		\hline 
	\end{tabular}
\end{table}

\begin{table}[t]
	\renewcommand\tabcolsep{2.6pt}%
	\centering%
	\caption{Ablation studies of proposed modules.} \label{Ablation}%
	\small
	\begin{tabular}{cccc|ccccc}
		\hline 
		
		Baseline & PCDF & CEU & SCA & Acc & MRE & MTE & $\rm mAP$ & $\rm recall$ \\ \hline
		\checkmark &  &  &  & 63.2 & 2.66 & 13.1 & 33.1 & 70.2 \\
		\checkmark & \checkmark &  &  & 60.0 & 2.92 & 6.02 & 41.7 & 78.8 \\
		\checkmark & \checkmark & \checkmark & & 70.1 & 2.02 & 5.99 & 52.1 & 90.5 \\
		\checkmark & \checkmark & \checkmark & \checkmark & \textbf{72.9} & \textbf{1.84} & \textbf{5.32} & \textbf{56.3} & \textbf{93.1} \\
		\hline 
	\end{tabular}
\end{table}

\noindent\textbf{Ablation Study - Tab. \ref{Ablation}.}
Finally, we perform ablation studies on SUN3D dataset to further analyze the effect of the proposed modules, including Progressive and Coherent Feature Drift (PCFD), Consensus Encoding Unit (CEU) and Spatial and Channel Attention (SCA) block. The 3DRegNet \cite{pais20203dregnet} is adopted as our baseline model. Since we have already proved that regression for $t$ and SVD for $R$ is the most suitable combination of estimation head, we use 3DRegNet with this alternative instead of the vanilla version.
We gradually add the proposed modules into the baseline model. First, we use the PCDF module to replace the CN Blocks \cite{moo2018learning} of 3DRegNet. The error of $t$ estimation significantly has decreased, which confirms the effectiveness of PCDF for regressing $t$. Then we adopt CEU to construct features for correspondence classification. As we can see, the classification accuracy is improved by 10\% compared with only using PCFD, leading to a better result of $R$. It shows that the proposed CEU can construct better classification features. Finally, we replace the CN blocks in 3DRegNet with SCA blocks, the performance of correspondences classification and $R$ estimation are further enhanced.

\section{Conclusion}
In this work, we develop a point cloud registration network named DetarNet, which decouples the estimation of rotation and translation. 
Specifically, we first propose a Progressive and Coherent Feature Drift (PCFD) module. It transforms the point cloud alignment process into a coherent drift operation in high-dimensional feature space and gradually estimates the translation. Then, we adopt a classification module to perform outlier pruning. It uses the proposed Consensus Encoding Unit (CEU) to construct feature for each correspondence, and adopts a Spatial and Channel Attention (SCA) for classification. Thus, the network can establish correct matches by taking advantages of the estimated $t$. Finally, $R$ matrix is obtained by performing weighted SVD. 
Extensive experiments on real scenes demonstrate the effectiveness of the proposed DetarNet.

\section{Acknowledgements}
This work was supported by the National Natural Science
Foundation of China under Grants 62176096 and 61991412.

\bibliography{egbib}
\nobibliography{aaai22}

\bigskip

\end{document}